\documentclass[11pt]{article}

\usepackage[margin=1in]{geometry}
\usepackage{times}
\usepackage{mathptmx}
\usepackage{amsmath, amssymb}
\usepackage{graphicx}
\usepackage{booktabs}
\usepackage{multirow}
\usepackage{hyperref}
\usepackage{url}
\usepackage{caption}
\usepackage{subcaption}
\usepackage{algcompatible}
\usepackage{algorithm}
\usepackage{algpseudocode}
\usepackage{longtable}
\usepackage{natbib}

\title{ButterflyMoE: Compression-Scalable Ternary Experts via Structured Butterfly Orbits}

\author{
  Aryan Karmore \\
  Indian Institute of Information Technology, Nagpur \\
  \texttt{bt24csd009@iiitn.ac.in}
}

\date{}

\begin{document}

\maketitle
\begin{abstract}
In current Mixture of Experts (MoE) architectures, linear memory scaling is present, the memory grows as the number of experts increases. $N$ independent expert weight matrices require $\mathcal{O}(N \cdot d^2)$ memory which exceeds the memory budget of edge devices. 
Current compression methods like quantization, pruning, and low-rank factorization reduce constant factors, but the scaling bottleneck is still unresolved. We introduce ButterflyMoE, a method which treats experts not as independent matrices but as geometric reorientations of a shared quantized substrate. Diversity amongst the experts arises from viewing different angles of the shared capacity and not from redundant storage. 
Learned rotations are applied to a shared ternary prototype. With this, each expert yields $\mathcal{O}(d^2 + N \cdot d \log d)$ memory-reducing per-expert cost from $\mathcal{O}(d^2)$ to $\mathcal{O}(d \log d)$. The key insight is that training these rotations with quantization reduces activation outliers and stabilizes extreme low-bit training where other static methods collapse. 
Across language modeling benchmarks, ButterflyMoE achieves 80$\times$ memory reduction at 8 experts with a highly favorable memory-accuracy tradeoff.At this 80x compression ButterflyMoE outperforms an equal memory dense baseline, showing that orbital parameterization extracts fundamentally more utility per byte. When scaled up to 256 experts, ButterflyMoE asymptotically compresses the memory by 150 $\times$. ButterflyMoE reduces the constant factor of linear scaling with compression ratio growing with the expert count. 
 
\end{abstract}

\section{Introduction}
 A conventional Mixture-of-Experts(MoE) layer with 64 experts and a model dimension of 256 stores millions of parameters. During training, additional memory is allocated for gradients, optimizer states, and activations, which increases the memory footprint. This makes deployment on resource-constrained devices challenging. This is the result of an architectural assumption that each expert requires separate parameters.

Current compression methods retain linear memory scaling. Quantization (QMoE \cite{frantar2023qmoe}, MoQE \cite{kim2023mixture}) reduces bit-width and achieves 4–5× compression but retains $O(N \cdot d^2)$ memory growth.Even 2-bit quantization requires 128 MB per layer at 256 experts. The structural question of whether $N$ experts need $O(N)$ separate parameter sets is not addressed.

In light of recent research on linear model connection, we propose that rather than being stored separately, experts can be formed from a shared prototype through straightforward transformations. This arrangement reduces per-expert memory from $\mathcal{O}(d^2)$ to $\mathcal{O}(d \log d)$, yielding a compression ratio that grows with expert count.

ButterflyMoE parameterizes $N$ experts as learned rotations of a single ternary-quantized substrate. Each expert $W_i$ is defined as:
\[
W_i = \mathcal{B}(\phi_i) \cdot W_{\text{base}} \cdot \mathcal{B}(\theta_i)^T,
\]
where $W_{\text{base}} \in \{-1, 0, +1\}^{d \times d}$ is a shared 1.58-bit weight matrix and $\mathcal{B}(\phi_i), \mathcal{B}(\theta_i)$ are expert-specific Butterfly matrices with $O(d \log d)$ parameters. Experts are never explicitly materialized. Inference proceeds by applying a rotation and a ternary matrix multiply. This yields $O(d^2 + N \cdot d \log d)$ memory.

This structural change has three benefits:(i) Memory compression improves with expert count achieving a 150x compression at 256 experts, (ii) Per-expert input rotations suppress activation outliers,(iii) A ternary substrate with rotations reduce quantization error by 97\% relative to post-training quantization.

ButterflyMoE achieves a highly favorable memory-accuracy tradeoff against dense and standard MoE baselines on WikiText-103, removing the primary memory bottleneck that currently prevents edge deployment.

\textbf{Contributions}
\begin{itemize}
    \item We introduce ButterflyMoE, combining ternary quantization with learned Butterfly rotations to achieve $O(d^2 + N \cdot d \log d)$ memory complexity.
    \item Theoretical roofline analysis shows ButterflyMoE has compute-bound execution on edge processors whereas Standard MoE architectures show memory-bound execution.
    \item A comprehensive empirical evaluation is done including accuracy memory tradeoffs, bitwidth accuracy ablation, comparison against rotation based quantization methods.

\end{itemize}

\section{Literature Review}
Mixture of Experts (MoE) models face a memory bottleneck during edge deployment. In standard MoE architectures, memory grows linearly with the number of experts. As the number of experts increases, memory usage grows.
A model with 64 experts and a hidden dimension of 512($d = 512$) stores 64 independent $512 \times 512$ weight matrices, which requires 256 MB. This exceeds the memory capacity of edge devices.

\subsection{Related Work}
Prior work is organized across 3 axes:MoE compression,Structured parameterization,Rotation-based quantization.
None of these current methods reduce the per-expert memory cost below $\mathcal{O}(d^2)$ while maintaining competitive accuracy.

\subsection{MoE Memory Compression}
Existing compression methods reduce compression factor but do not break the linear memory scaling. Decomposition methods like MoBE \cite{chen2025mobe} represent experts as a linear combination of $K$ shared bases which yields $\mathcal{O}(K \cdot d^2 + N \cdot K)$ memory consumption. 

Low-rank factorization\cite{yang2024moe} and expert pruning-merging \cite{zhao2025puzzlemoe} methods, like Puzzle MoE, achieve $2\times$  compression but cannot scale beyond 128 experts on edge devices. A common limitation of these architectures is that the expert identities are irreducible. This forces the explicit materialization of all expert weight matrices.

\subsection{Structured and Rotation-Based Quantization}

Prior work shows orthogonal rotations can suppress activation outliers before quantization using the invariance property $y = Wx = (WQ^T)(Qx)$. QuaRot \cite{ashkboos2024quarot} applies Hadamard transformations,  SpinQuant \cite{liu2025spinquant} optimizes full $n \times n$ rotation matrices requiring $\mathcal{O}(n^2)$ parameters. 

These methods apply a single global rotation per layer for post-training quantization. A key assumption is made that outlier distributions remain stable during training, but this assumption is violated by channel drift and cannot capture per-expert specialization.

Concurrently, structured matrices like Monarch \cite{dao2022monarch} and Butterfly \cite{dao2019learning} parameterize orthogonal transformations with $\mathcal{O}(d \log d)$ parameters. Prior work uses them as a drop-in replacement for dense layers and not for parameterizing expert identity.

ButterflyMoE bridges these lines of work. Instead of applying a global transformation post-training, ButterflyMoE embeds distinct butterfly rotations into the expert parameterization during training. Each expert's rotation is jointly optimized with a shared ternary substrate to specialize its feature subspace. Butterfly matrices are chosen over Monarch matrices because they guarantee orthogonality by construction, which is essential for outlier suppression. Monarch matrices, which are block diagonal, require additional constraints to preserve this property \cite{dao2022monarch}.

\section{Methodology}
\subsection{Problem Formulation and Edge Constraints}
A MoE layer routes inputs by a gating network $g$ to top-$k$ experts: 

 $y = \sum_{i \in \text{TopK}(g(x))} g_i(x) W_i x$.

Storing $N_E$ independent matrices requires $M_{\text{MoE}} = N_E \cdot d_{\text{ff}} \cdot d_{\text{model}} \cdot b_{\text{precision}}$ bytes. 
This $\mathcal{O}(N_E \cdot d^2)$ scaling causes two critical failures on edge devices: a \textbf{memory wall} (e.g., 64 experts at $d=512$ require 256 MB, exceeding typical RAM budgets) and a \textbf{bandwidth bottleneck} (repeatedly loading these weights from DRAM consumes $\sim$13 mJ per pass, unsuitable for battery-powered devices). 
The core objective is to find a parameterization which satisfies: (1) per-expert memory cost of $\mathcal{O}(d \log d)$ instead of $\mathcal{O}(d^2)$, (2) preserved expert diversity, and (3) accuracy preservation.

\subsection{ Core Insight: Experts as Orbits of a Quantized Prototype}
Recent studies on linear mode connectivity \cite{garipov2018loss,ainsworth2022git} show that neural networks trained on similar data lie on low dimensional manifolds through group transformations such as permutations of neurons, feature space rotations or rescalings.

Instead of storing $N_E$ independent matrices  $\{W_i\}$, they are represented as orbital variations of a single quantized prototype $W_{\text{base}}$ under learnable orthogonal transformations. This is given in the mathematical formulation:

\subsection*{Mathematical Formulation}
Define each expert as:
\begin{equation}
    W_i = \mathcal{R}_i^{\text{out}} \cdot W_{\text{base}} \cdot (\mathcal{R}_i^{\text{in}})^T
\end{equation}
where:
\begin{itemize}
    \item $W_{\text{base}} \in \{-1, 0, +1\}^{d_{\text{ff}} \times d_{\text{model}}}$ is a ternary-quantized substrate (1.58 bits/weight).
    \item $\mathcal{R}_i^{\text{in}}, \mathcal{R}_i^{\text{out}} \in O(d)$ are expert-specific orthogonal rotations.
\end{itemize}

This works because of the following reasons:
\begin{itemize}
    \item \textbf{Shared capacity:} $W_{\text{base}}$ captures universal features such as syntax, basic semantics
    \item \textbf{Orbital diversity:} The rotations $\mathcal{R}_i$ create different views of the shared parameter substrate,allowing each expert $i$ to specialize by aligning $W_{\text{base}}$ with domain-specific feature subspaces.
    \item \textbf{Quantization resilience:} Learned input rotations $\mathcal{R}_i^{\text{in}}$ suppress activation outliers.
\end{itemize}

Storing $W_{\text{base}}$ requires $O(d^2)$ bits (quantized), while rotations need only $O(\log d)$ parameters when structured effectively using Butterfly matrices.

\subsection{Method Overview}
\begin{figure}[h]
    \centering
    \includegraphics[width=0.3\textwidth]{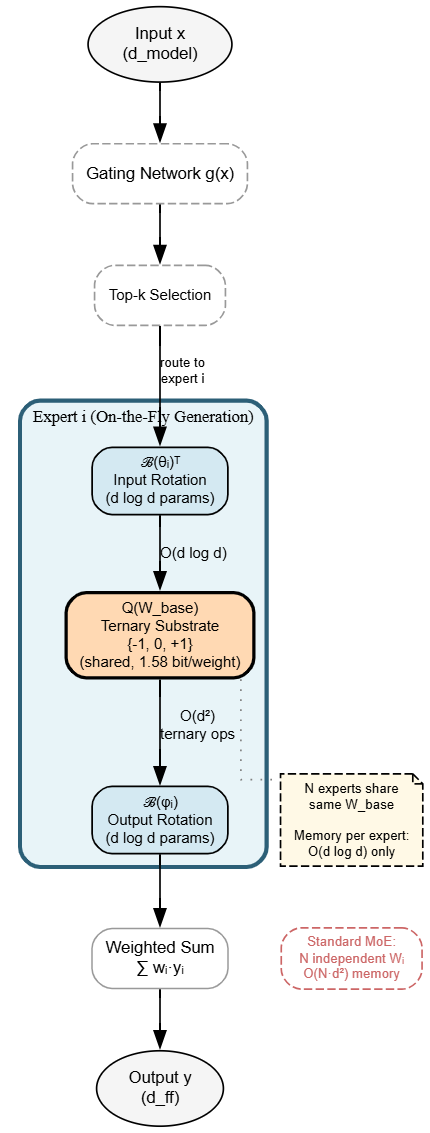}
    \caption{Top-$k$ gating instantiates experts via lightweight rotations of a shared ternary base matrix $\mathbf{W}_{\text{base}}$ achieving $\mathcal{O}(d \log d)$ parameters per expert and weighted sum aggregation.}
    \label{fig:method_overview}
\end{figure}
Figure ~\ref{fig:method_overview} shows ButterflyMoE replacing independent weight matrices with sequential operations: An input rotation, A shared ternary matrix multiplication, An output rotation.
The model is trained with standard cross-entropy loss and load balancing.

\subsection{ Parameterization via Butterfly Matrices}

Orthogonal transformations $\mathcal{R} \in O(d)$ require $O(d^2)$ parameters and therefore offer no savings in memory footprint.
This can be solved using butterfly matrices \cite{dao2019learning} where rotations are parameterized with only $O(d \log d)$ scalars through a recursive block diagonal decomposition
This is defined below:

A Butterfly matrix $\mathcal{B}(\theta) \in \mathbb{R}^{d \times d}$ with $d = 2^m$ is:
\begin{equation}
    \mathcal{B}(\theta) = \prod_{\ell=1}^{m} \mathcal{D}_\ell(\theta) \mathcal{P}_\ell
\end{equation}
where:
\begin{itemize}
    \item $\mathcal{D}_\ell(\theta)$ is block-diagonal with $2^\ell \times 2^\ell$ blocks, each applying a 2D Givens rotation:
    \begin{equation}
        \begin{bmatrix}
            \cos \alpha & -\sin \alpha \\
            \sin \alpha & \cos \alpha
        \end{bmatrix}
    \end{equation}
    \item $\mathcal{P}_\ell$ is a fixed permutation (perfect shuffle).
    \item $\theta = \{\alpha_{\ell,j}\}_{\ell=1,\ldots,m; j=1,\ldots,d/2}$ are learnable angles.
\end{itemize}

\subsection*{Complexity}
\begin{itemize}
    \item \textbf{Storage:} $O(d \log d)$ parameters so for $d=512$, $512 \log_2 512 = 4608$ angles.
    \item \textbf{Computation:} $O(d \log d)$ FLOPs per forward pass.
    \item \textbf{Expressivity:} Monarch matrices \cite{dao2022monarch} shows that $O(\log d)$ butterfly layers approximate a broad class of orthogonal transformations, with vanishing error as depth increases.
\end{itemize}

For a more practical implementation $\log_2 d$ layers are used, each parameterized by $d/2$ Givens rotations $\alpha \in [-\pi, \pi]$ optimized by standard gradient based methods.

\subsubsection{Non-Power-of-Two Dimensions}
Standard Butterfly factorization requires $d = 2^m$ for integer $m$, this excludes common LLM dimensions such as 768 and 5120. ButterflyMoE handles arbitrary dimensions via zero padding.

For a dimension $d$ that is not a power of 2, we pad the input to $\hat{d} = 2^{\lceil \log_2 d \rceil}$ before applying the Butterfly transformation and truncate the output back to $d$ afterwards:
\[
\hat{x} = \text{pad}(x, \hat{d}), \quad \hat{y} = \mathcal{B}(\theta)\hat{x}, \quad y = \hat{y}_{[1:d]}
\]
. The padding introduces at most a 2$\times$ overhead in the butterfly-like computation since $\hat{d} < 2d$, however the orthogonality remains preserved in the active dimensional subspace. The padded dimensions are zeroed out before and discarded after the transformation, leaving the rotation structure intact.
\subsection{Quantized Substrate and Outlier Suppression}

The shared substrate $W_{\text{base}}$ is quantized to $\{-1, 0, +1\}$ (1.58 bits/weight) via AbsMean scaling \cite{ma2024era}:
\begin{equation}
    Q(W) = \gamma \cdot \text{round}\left(\frac{W}{\gamma}\right), \quad \gamma = \frac{1}{d_{\text{ff}} d_{\text{model}}} \sum_{ij} |W_{ij}|
\end{equation}
where $\text{round}(\cdot)$ clips to the ternary set. Gradients pass through using the Straight-Through Estimator ($\partial Q/\partial W = I$) \cite{bengio2013estimating}. 
Clipping or per-channel scaling are methods used for outlier suppression, but they destroy information and can fail to distribute energy across dimensions \cite{ashkboos2024quarot}. ButterflyMoE uses a different approach, instead, each expert's input rotation $\mathcal{B}(\theta_i)$  is trained jointly with shared substrate $W_{\text{base}}$. The gradients $\partial \mathcal{L}/\partial \theta_i$, which dynamically rotate the activations into quantization-friendly bases that align with the frequent patterns in the ternary grid to suppress the outliers.

\subsection{Training Objective and Initialization}
We optimize the model end-to-end using a standard cross-entropy loss combined with a load-balancing loss \cite{fedus2022switch}:
\begin{equation}
    \mathcal{L} = \mathcal{L}_{\text{CE}}(y, \hat{y}) + \lambda_{\text{balance}} \sum_{i=1}^{N_E} \left( \frac{n_i}{N_{\text{total}}} - \frac{1}{N_E} \right)^2
\end{equation}
where $n_i$ is the number of tokens routed to expert $i$ and $\lambda_{\text{balance}} = 0.01$. To prevent expert collapse \cite{chi2022representation}, the butterfly angles $\{\theta_i, \phi_i\}$ are independently initialized from $\mathcal{N}(0, 0.01^2)$. This breaks symmetry and promotes specialization from the start of training


\subsection{Analytical Properties}
Three analytical properties define the memory, quantization, and computational behavior of ButterflyMoE. First, the exact memory scaling of the proposed orbital parameterization is derived. Then, an asymptotic compression ratio is established relative to standard MoE. Lastly, we describe the inference-time computational complexity under top-$k$ routing.

\textbf[Memory Scaling]
For $N_E$ experts with dimensions $d_{\text{model}}, d_{\text{ff}}$, ButterflyMoE memory is:
\begin{equation}
M_{} = \frac{1.58}{8} d_{\text{ff}} d_{\text{model}} + N_E \cdot \left(\frac{d_{\text{model}}}{2} \log_2 d_{\text{model}} + \frac{d_{\text{ff}}}{2} \log_2 d_{\text{ff}}\right) \cdot 2
\end{equation}

\textbf[Asymptotic Compression]
As $N_E \to \infty$ with fixed $d_{\text{model}}$ and $d_{\text{ff}}$, the substrate cost $\mathcal{O}(d_{\text{model}} \cdot d_{\text{ff}})$ amortizes, and per-expert memory dominates. The asymptotic compression ratio is:
\begin{equation}
\lim_{N_E \to \infty} \frac{M_{\text{Standard MoE}}}{M_{\text{ButterflyMoE}}} 
= \frac{d_{\text{model}} \cdot d_{\text{ff}} \cdot 4}{\left(\frac{d_{\text{model}}}{2} \log_2 d_{\text{model}} + \frac{d_{\text{ff}}}{2} \log_2 d_{\text{ff}}\right) \cdot 2 }
\end{equation}

For $d_{\text{model}} = 512$ and $d_{\text{ff}} = 2048$:
\begin{align*}
&= \frac{512 \cdot 2048 \cdot 4}{(256 \cdot 9 + 1024 \cdot 11) \cdot 2} \\
&= \frac{4{,}194{,}304}{27{,}136} \\
&\approx 154.5\times
\end{align*}

At $N_E{=}8$, the substrate term is still not negligible, which yields an $80\times$ memory reduction. As the number of experts increases to 256, the per-expert term dominates, and the compression approaches the asymptotic limit, yielding the observed $150\times$. 

As $N_E$ increases, compression ratio increases demonstrating that ButterflyMoE's advantage grows with scale.

\textbf[Computational Complexity]
Inference FLOPs per token with top-$k$ routing:
\begin{equation}
C_{\text{ButterflyMoE}} = O(k \cdot d \log d) + O(d^2) \quad \text{(ternary multiply)}
\end{equation}
Here ternary multiply involves only additions(no multiplications), achieving $\sim$10$\times$ lower energy per operation.

\section{Experiments}
\subsection{Experimental Setup}
All experiments are conducted on WikiText-103 using the GPT-2 tokenizer with a vocabulary size of 50,257. The model features a 4-layer architecture with a model dimension of 512, an FFN dimension of 2048, 8 experts (top-2 routing), 8 attention heads, and a sequence length of 128. 

No recovery stages or post-training procedures were required as the model trains stably end-to-end from random initialization under standard cross entropy loss with load balancing.
The main language modelling experiment is trained for more than 15000 steps to ensure full convergence, whereas subsequent ablations use a fixed budget of 15–20 epochs to isolate architectural differences.

\subsection{Language Modeling Performance}
\subsubsection{Language Modeling Performance on Wikitext-103}

ButterflyMoE achieves a PPL of 81.19 at epoch 40, utilizing only 0.40 MB of memory, which is an 80x memory reduction. This memory reflects only FFN/Expert weights per layer. The attention, embeddings, and layer norms are kept identical across configurations and are hence omitted. Standard MoE and dense FFN converge by epoch 30 while ButterflyMoE requires roughly 1.8x longer training due to joint optimization of the shared ternary substrates and per-expert rotations~\cite{ma2024era}. 

In the case of standard MoE, independent experts are optimized freely from random initialization, which enables convergence. In ButterflyMoE the shared substrate must learn all of the universal features which are useful to all experts while each expert's rotations orient this capacity towards specialized subspaces. This joint optimization is much more complex but uses substantially less memory, with ButterflyMoE using 0.40 MB versus 32 MB for standard MoE. The saved memory can be reinvested into embeddings or additional layers, likely surpassing standard MoE on an equal memory budget

\begin{table}[t]
\centering
\caption{Validation perplexity on WikiText-103. \textbf{All memory values represent the FFN/expert weights only, per layer, in FP32 (or equivalent bit-width)}.}
\label{tab:perplexity}
\begin{tabular}{lcccc}
\toprule
\textbf{Method} & \textbf{FFN Config} & \textbf{Val PPL} & \textbf{FFN Mem} & \textbf{Compr.} \\
 & ($d_{model} \times d_{ff}$) & & \textbf{(MB)} & \\
\midrule
Standard MoE & $512 \times 2048$ (8 exp.) & 71.75 & 32.00 & 1$\times$ \\
Dense FFN & $512 \times 2048$ & 83.01 & 8.00 & 4$\times$ \\
\midrule
Dense FFN (Matched Mem.) & $512 \times 128$ &135.46  & 0.50 & 64$\times$ \\
\textbf{ButterflyMoE} & $512 \times 2048$ (8 exp.) & \textbf{81.19} & \textbf{0.40} & \textbf{80$\times$} \\
\bottomrule
\end{tabular}
\end{table}
To isolate the effects of orbital parametrization from parameter reduction, ButterflyMoE is compared against an equal-memory dense baseline. This baseline uses a little more memory than ButterflyMoE and operates in full FP32 without ternary quantization. Both of these baselines are trained for identical steps. Despite these advantages the dense baseline suffers severe capacity collapse while ButterflyMoE achieves 81 PPL. This shows that sharing a quantized substrate among orbital experts extracts more representational quality than simply shrinking a dense network.

\begin{figure}[h]
    \centering
    \includegraphics[width=0.4\linewidth]{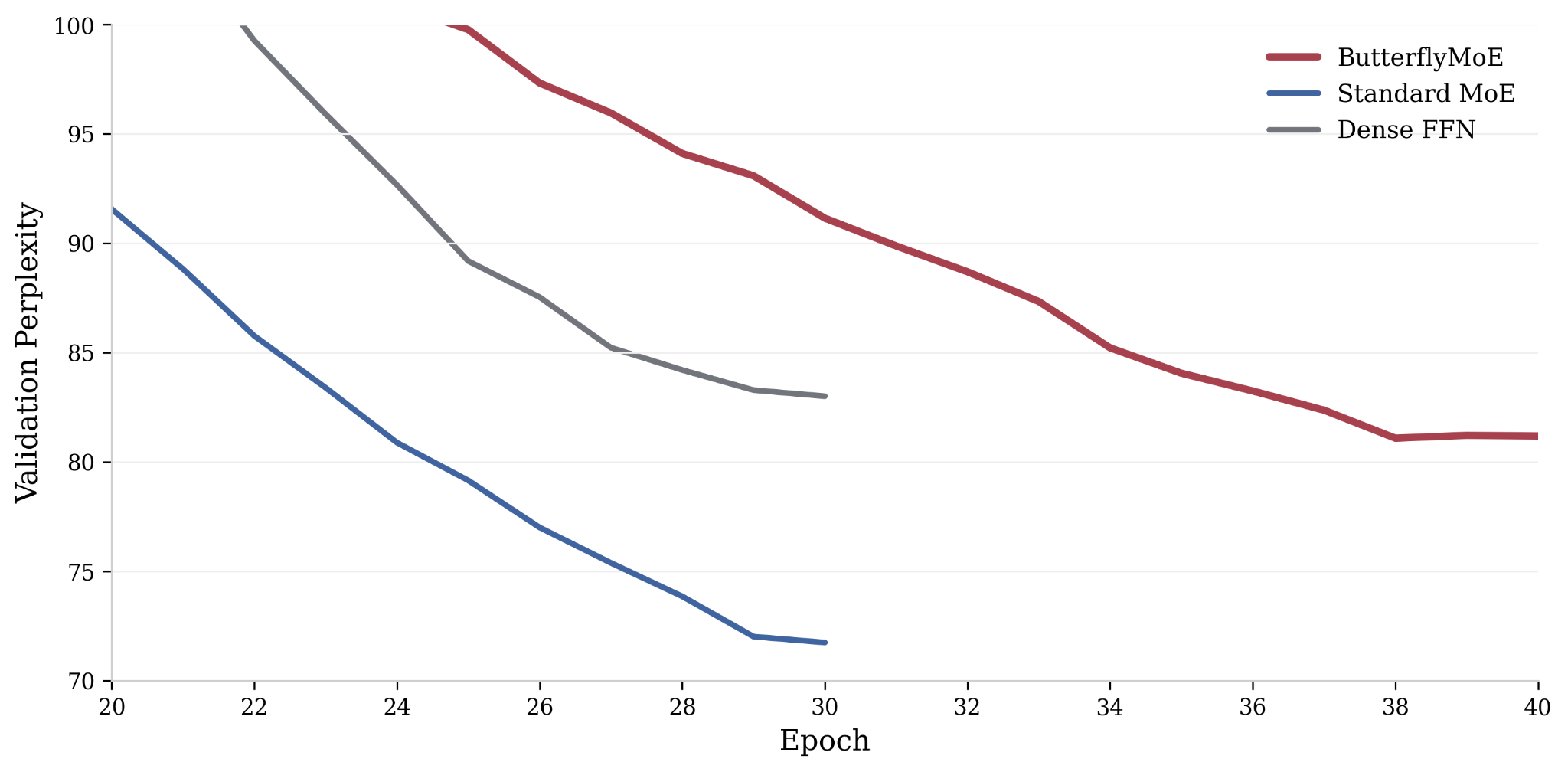}
    \caption{ Validation perplexity during the late stages of training (epochs 20–40) on the WikiText-103 benchmark.}
    \label{fig:zoomed}
\end{figure}

\subsubsection{Scaling of Experts}

The number of experts is increased to 16 and the top-k value is increased from 2 to 4 to keep a fixed routing ratio of 25\%. Standard MoE shows a small improvement and ButterflyMoE decouples expert count from the memory-accuracy tradeoff. This is expected as in standard MoE architectures accuracy gains from scaling experts arise from increasing routing ratio. The critical difference is that Standard MoE pays an additional 32 MB for these experts whereas ButterflyMoE only pays 0.21 MB. ButterflyMoE decouples expert count from the memory-accuracy tradeoff.

\begin{table}[t]
\centering
\caption{Scaling validation at a fixed 25\% routing ratio (per layer, FP32).(Reported at 11,000 steps to isolate scaling dynamics under a fixed computational budget)}
\label{tab:scaling_validation}
\begin{tabular}{lccccc}
\toprule
\textbf{Method} & \textbf{Experts} & \textbf{Routing} & \textbf{Val PPL} & \textbf{Mem (MB)} & $\boldsymbol{\Delta}$ \textbf{PPL} \\
\midrule
Standard MoE & 8  & Top-2 (25\%) & 118.47 & 32.00 & - \\
Standard MoE & 16 & Top-4 (25\%) & 109.64 & 64.00 & {\textbf{-10.0}} \\
\midrule
\textbf{ButterflyMoE} & 8  & Top-2 (25\%) & \textbf{128.32} & \textbf{0.40} & - \\
\textbf{ButterflyMoE} & 16 & Top-4 (25\%) & \textbf{128.38} & \textbf{0.61} & {\textbf{+0.06}} \\
\bottomrule
\end{tabular}
\end{table}

As shown in the table ~\ref{tab:scaling_validation}, the validation PPL is identical across both configurations for ButterflyMoE. This shows that adding more orbital experts does not introduce new parameters but provides a dense sampling of orthogonal transformations around the fixed Shared Substrate. As the representational capacity is strictly bounded by the substrate, the PPL remains dictated by the routing ratio and the quality of the shared substrate, not the expert count.
It suggests that scaling to massive expert counts in ButterflyMoE could provide fine-grained routing granularity without the associated memory explosion. Validation perplexity at larger expert counts remains as future work, and this experiment shows that ButterflyMoE with 16 experts trains stably.

Standard MoE requires a linear memory tax to increase the routing granularity. ButterflyMoE decouples these dependencies as experts essentially become free parameters. This suggests that in production ButterflyMoE could utilize massive expert counts purely to improve load balancing and reduce per-expert latency, without suffering the catastrophic memory penalties of standard MoE. 

\subsubsection{Non-Power-of-Two Dimensions.} 
\begin{figure}[h]
    \centering
    \includegraphics[width=0.4\linewidth]{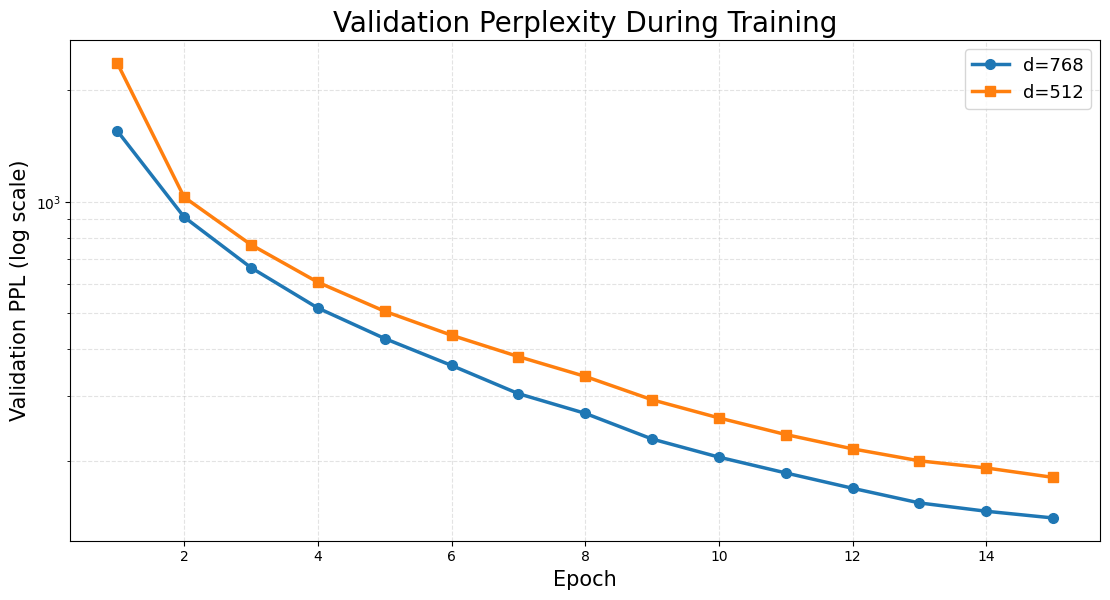}
    \caption{
        Ablation of non power of two dimension size in ButterflyMoE. Scaling the model dimension from 512 to 768 results in a substantial reduction in validation perplexity. Reported at 15 epochs.
    }
    \label{fig:d=768}
\end{figure}

ButterflyMoE trains stably at $d=768$ with loss reduction across all 15 epochs. This confirms that the zero padding approach generalizes beyond power-of-two dimensions and that it can work without architectural modification or accuracy pathology. Fig ~\ref{fig:d=768} shows the validation perplexity between $d=768$ and $d=512$.

Orthogonality is preserved on the full padded dimension $\hat{d}$; restricting to the active $d$ coordinates yields an approximately orthogonal transformation, with deviation governed by the padding ratio $\hat{d}/d$, which we found empirically sufficient for stable outlier suppression.

\subsubsection{Comparison with different bitwidth.}  
\begin{table}[h]
\centering
\caption{Validation Perplexity of 1 bit, 1.58 bit, and 2 bit at 15 epochs. Reported at 15 epochs, 7,500 steps.}
\label{tab:bit}
\begin{tabular}{lccc}
\toprule
\textbf{Method} & \textbf{Value} \\
\midrule
1 bit & 205.59 \\
1.58 bit & 180.67 \\
2 bit & 189.61 \\
\bottomrule
\end{tabular}
\end{table}

A bitwise ablation shows that ternary quantization outperforms both binary and two-bit quantization. Binary quantization collapses due to insufficient representational capacity. Two-bit quantization suffers from increased gradient noise in the STE, as the additional quantization level introduces discontinuities that destabilize training and increase the memory cost. Ternary strikes an optimal balance with sufficient expressivity at three levels, stable gradients via coarse binning, and natural sparsity through the zero value. Final perplexity scores are shown in table ~\ref{tab:bit} at 15 epochs.

\subsection{Comparison with Rotation-Based Quantization.} 
SpinQuant~\cite{liu2025spinquant} applies learned rotation to model activations prior to quantization. This redistributes outlier magnitudes across channels so that activations can be quantized to low bit widths with minimal accuracy loss.
SpinQuant style ablation achieves a validation PPL of 132.54 against ButterflyMoE's 143.58 at epoch 20 while consuming 3.4$\times$  more memory. SpinQuant's single-layer FFN weights ($w_1 + w_2$) contain twice as many parameters as ButterflyMoE's entire shared substrate, shared across all eight experts. SpinQuant style ablation follows its original dense formulation with no expert routing, whereas ButterflyMoE distributes capacity across eight experts. The residual PPL gap is consistent with the known tendency for dense architectures to outperform MoE at small parameter and data budgets. Both models continue to descend after 20 epochs, showing that none of the models have converged.

\subsubsection{Comparison with Monarch Matrices.}

Monarch MoE achieves a validation PPL of 152.72 at epoch 20 compared to ButterflyMoE's 143.58 while consuming 1.72 times more memory. This gap is due to Monarch's block-diagonal parameterization, which requires more parameters per expert at practical block sizes. Secondly Monarch matrices do not preserve orthogonality by construction, which is essential for the outlier suppression mechanism in ButterflyMoE. While Monarch matrices offer better hardware utilization through their batch matrix multiplication in principle, the orthogonal guarantee of Givens angle rotation provides a structural advantage for ternary quantization that Monarch matrices cannot replicate without additional constraints ~\cite{dao2022monarch}.
\begin{table}[h]
\centering
\caption{ButterflyMoE vs MonarchMoE at epoch 20 ($d=512$, $N_E=8$). Reported at 20 epochs}
\begin{tabular}{lccc}
\toprule
\textbf{Method} & \textbf{Val PPL} \\
\midrule
MonarchMoE & 152.72 \\
ButterflyMoE & 143.58 \\
\bottomrule
\end{tabular}
\end{table}

\subsection{Memory Scaling and Edge Deployability}
\subsubsection{Comparison with MoE compression Methods.}

ButterflyMoE is compared against a dense FFN with matched parameter count and a standard MoE with independently parameterized experts.
ButterflyMoE reparameterizes experts as structural transformations on a single substrate, thus eliminating the need to save separate expert matrices. This results in significant compression factors as shown in Table-1. Comparison with low rank adaptors are not made as they do not scale expert count.

\begin{table*}[t]
\centering
\caption{Comparison of MoE compression methods for models with 64 experts ($d=512$, $d_{ff}=2048$). The memory of ButterflyMoE is calculated theoretically for 64 experts.}
\label{tab:comprehensive_comparison}
\setlength{\tabcolsep}{8pt}
\begin{tabular}{lccc}
\toprule
\textbf{Method} & \textbf{Memory} & \textbf{Compression} & \textbf{Edge} \\
 & \textbf{Scaling} & \textbf{Ratio (64)} & \textbf{Deployment} \\
\midrule
Standard MoE & $O(N \cdot d^2)$ & 1.0$\times$ & 256 MB \\
QMoE\cite{frantar2023qmoe} & $O(N \cdot d^2)$ & 10-20$\times$ & 13-26 MB \\
MoQE (2-bit)\cite{kim2023mixture} & $O(N \cdot d^2)$ & 5.0$\times$ & 51 MB \\
PuzzleMoE\cite{zhao2025puzzlemoe} & $O(N \cdot d^2)$ reduced & 2$\times$ & 128 MB \\
MC\cite{huang2024mixture} & $O(N \cdot d^2)$ reduced & 4.0$\times$ & 64 MB \\
ButterflyMoE & $O(d^2 + N \cdot d \cdot \log d)$ & \textbf{138$\times$} & \textbf{1.9 MB} \\
\bottomrule
\end{tabular}
\end{table*}
\vspace{0.5cm}

\label{tab:moe_comparison}

QMoE, MoQE, and MoBE report accuracy on C4, WMT Translation, and a 15-benchmark suite on production-scale LLMs respectively. None of these methods report Wikitext 103 perplexity, making direct numerical comparison on this metric infeasible without re-implementing each method end-to-end in our training pipeline. Direct comparison is not feasible given these methods are closed source and the full code is not available. Re-implementing these methods as MoE architectures would require making arbitrary and unverified design choices, which would introduce confounding variables. The memory ratios of QMoE, MoQE, and MoBE are theoretical scaling bounds derived from their published architectures.

\subsubsection{Memory Consumption as a function of the number of experts.}

ButterflyMoE scales to 150$\times$ compression at 256 experts ($d{=}512$), 
requiring only 4.70~MB compared to standard MoE's 1024~MB. The compression ratio improves as $N_E$ rises. Table ~\ref{tab:experts_metrics} shows the detailed per expert-memory comparison with ButterflyMoE showing significantly lesser memory as the number of experts increases.

\begin{table}[h]
\centering
\begin{tabular}{cccc}
\toprule
\textbf{Experts} & \textbf{Standard MoE} & \textbf{ButterflyMoE} & \textbf{Compression} \\
\midrule
2               & 8.00                   & 0.25            & 32.0                \\
4               & 16.00                  & 0.30            & 53.15                \\
8               & 32.00                  & 0.40            & 80.0                \\
16              & 64.00                  & 0.61            & 104.65               \\
32              & 128.00                 & 1.03            & 124.80               \\
64              & 256.00                 & 1.85            & 138.10               \\
128             & 512.00                 & 3.51            & 145.87               \\
256             & 1024.00                & 6.82            & 150.09               \\
\bottomrule
\end{tabular}
\caption{Performance and compression metrics for varying numbers of experts in Standard MoE and ButterflyMoE models.}
\label{tab:experts_metrics}
\end{table}
\subsection{Quantization Stability via Learned Rotations}
\begin{figure}[t]
    \centering
    \includegraphics[width=0.4\linewidth]{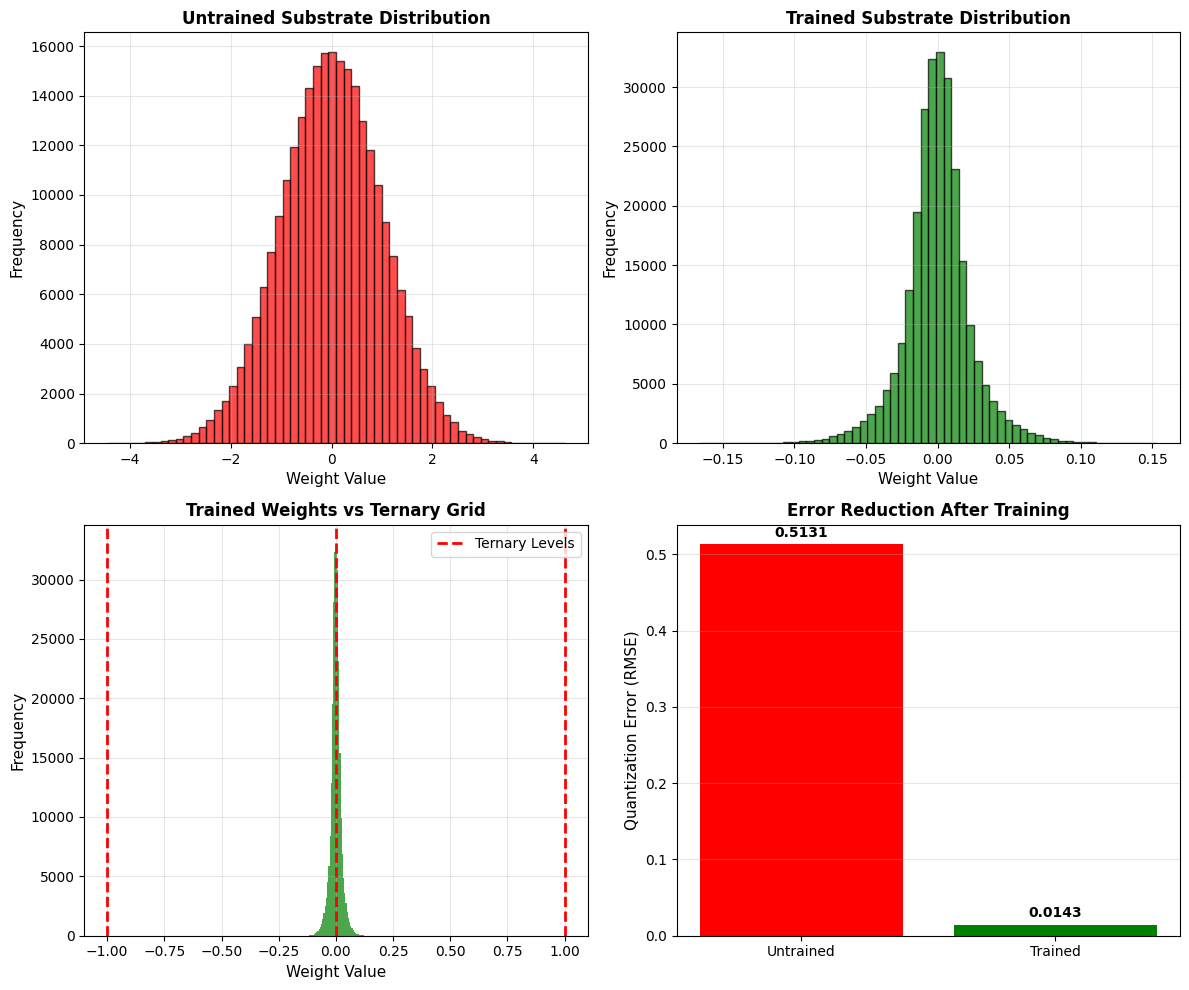}
    \caption{
        \textbf{Quantization stability via learned rotations.} 
        (Top left)~Untrained substrate weight distribution spreads across 
        $[-4, +4]$. (Top right)~After training, weights cluster near ternary 
        grid $\{-1, 0, +1\}$. (Bottom left)~Trained weights (green) align 
        with quantization levels (red dashed lines). (Bottom right)~Mean 
        squared error: 51.3\% (untrained) → 1.43\% (trained), a 97.2\% reduction.
    }
    \label{fig:quantization}
\end{figure}

Activation outliers dominate quantization error, so extreme low-bit quantization traditionally fails \cite{xiao2023smoothquant}. Our theoretical mechanism shows that per-expert rotations suppress these activation outliers. Figure~\ref{fig:quantization} demonstrates the empirical result of the joint optimization. This shows the dynamic alignment of the shared weight substrate with the quantization grid.

The weight quantization mean squared error drops from 0.513 in the untrained substrate to 0.0143 after training. The untrained distribution spreads across the range (-4, 4), but the trained distribution is tightly concentrated across the ternary grid. This concentration shows that the learned rotations are making quantization stable and thus eliminating the need for clipping and post-training recovery stages

\subsection{Expert Specialization}
We measure expert diversity via pairwise cosine similarity between expert outputs on the WikiText-103 validation set. Formally, for $N_E$ experts, let
\begin{equation}
S_{ij} = \frac{\langle \mathbf{y}_i, \mathbf{y}_j \rangle}{\|\mathbf{y}_i\| \|\mathbf{y}_j\|},
\end{equation}
where $\mathbf{y}_i \in \mathbb{R}^{B \times L \times d_{\text{ff}}}$ is the output of expert $i$ flattened over tokens. The diversity score is
\begin{equation}
D = 1 - \frac{2}{N_E (N_E - 1)} \sum_{i < j} S_{ij},
\end{equation}
ranging from 0 (complete collapse) to 1 (orthogonal outputs).
Experts maintain distinct behaviour despite operating a shared quantized substrate.Off diagonal values span 0.08-0.14 with multiple expert pairs below 0.1. This indicates that the learned rotations induce meaningfully differentiated feature spaces.

Despite a $\sim$150$\times$ decrease in parameter storage, ButterflyMoE gets a diversity score of 0.87 as opposed to 0.912 for a Standard MoE which is a 5\% difference.

The small performance gap shows a regularization effect and the low diagonal similarities show that the experts remain distinct.

\subsection{Throughput and Latency Analysis}

\begin{table}[h]
\centering
\caption{Single-token latency ($\mu$s, batch size 1) and inference throughput 
(tok/s, batch size 16) on a Tesla T4 GPU. }
\label{tab:throughput}
\begin{tabular}{lcccc}
\toprule
\textbf{Method} & \textbf{Params} & \textbf{FLOPs/token} & 
\textbf{Latency ($\mu$s)} & \textbf{Throughput (tok/s)} \\
\midrule
Standard MoE           & 33.6M & 16.8M & 1,209          & 30,509 \\
ButterflyMoE (PyTorch) & 2.3M  & 4.3M  & 10,921         & 11,778 \\
ButterflyMoE (Triton)  & 2.3M  & 4.3M  & \textbf{1,298} & \textbf{29,476} \\
\bottomrule
\end{tabular}
\end{table}

Table~\ref{tab:throughput} shows inference latency and throughput on a 
Tesla T4 GPU. The native PyTorch implementation of ButterflyMoE results in a 9.03x latency increase compared to Standard MoE 
This is due to two structural inefficiencies: $\log_2 d$ sequential kernel launches per rotation 
and $k$ separate cuBLAS calls against the shared 
$W_{\text{base}}$ matrix.

First all $\log_2 d$ butterfly layers are fused into a single kernel launch per rotation. The full $d$ dimensional token row is loaded into registers once and all layers execute in register with \texttt{tl.static\_range} , which is unrolled at compile time. The result is written back in a single store. No intermediate values touch HBM between the layers.
Second the active experts' inputs are stacked into a $(k \times d_{\text{model}})$ matrix and dispatched as one GEMM against this. This replaces $k$ matrix multiplications with a single call that cuBLAS schedules efficiently.

These optimizations reduce ButterflyMoE latency from 10,921\,$\mu$s 
to 1,298\,$\mu$s which is 8.41$\times$ speedup over the PyTorch baseline and recovers 
throughput from 11,778 to 29,476 tok/s, within 
3.4\% of Standard MoE (30,509 tok/s). This residual gap falls within run-to-run variance which is std.\,$\approx$\,130\,$\mu$s across 500 runs. ButterflyMoE achieves $14\times$ parameter reduction and $4\times$ FLOP reduction at no wall-clock cost relative to Standard MoE.

\section{Roofline Model Analysis}
The memory reduction and compression achieved by ButterflyMoE alter the arithmetic intensity of the MoE inference. ButterflyMoE shifts the execution regime from memory-bound to compute-bound on resource-constrained edge processors. This behavior is analyzed through a roofline performance model using the ARM Cortex-A72 as a representative edge platform—a quad-core 1.5 GHz processor with 128-bit NEON SIMD units. This is widely deployed in devices such as Raspberry Pi 4 and various other embedded controllers.

\subsection{Hardware Platform Characterization}
The Cortex-A72 achieves a theoretical peak FP32 throughput of 48 GFLOP/s with 4-wide NEON FMA operations (8 FLOPs/cycle/core $\times$ 4 cores $\times$ 1.5 GHz). Memory bandwidth is limited by a single-channel LPDDR4 interface. STREAM-like benchmarks report 6 GB per second in practice. The ridge point is therefore :

\begin{equation}
\text{Ridge Point} = \frac{\beta}{\pi} = \frac{6 \text{ GB/s}}{48 \text{ GFLOP/s}} = 8.0 \text{ FLOP/Byte}
\end{equation}

Kernels with the arithmetic intensity below 8 FLOP/byte are memory-bound and those above 8 FLOP/byte are compute-bound.

\subsection{Operation Analysis from Algorithm 1}
For our configuration ($d_{\text{model}} = 512$, $d_{\text{ff}} = 2048$, $N_E = 8$, top-$k=2$), we derive the per-token costs directly from the forward pass described in Section 3.3. 

\textbf{FLOPs:} The total per-expert computation is 1.13 MFLOPs, dominated by 1.05M additions required for the ternary matrix multiplication. The input and output butterfly rotations contribute 81.4K flops combined. For top-$k=2$, this is 2.26 MFLOPs/token. This is computed as shown below
\begin{itemize}
    \item \textbf{Input butterfly rotation} $B(\theta_i)^T x$: The butterfly matrix comprises $\log_2 d_{\text{model}} = 9$ layers, each with $d_{\text{model}}/2 = 256$ Givens rotations. Each $2 \times 2$ rotation requires 6 FLOPs (4 multiplies + 2 additions), yielding $9 \times 256 \times 6 = 13,824$ FLOPs.
    \item \textbf{Ternary matrix multiply} $Q(W_{\text{base}}) \cdot x_{\text{rot}}$: The substrate $W_{\text{base}} \in \{-1, 0, +1\}^{2048 \times 512}$ requires only additions (no multiplications) since ternary weights are resolved to $\pm1$ or 0. This yields $d_{\text{ff}} \times d_{\text{model}} = 1,048,576$ additions.
    \item \textbf{Output butterfly rotation} $B(\phi_i) y_{\text{base}}$: With $\log_2 d_{\text{ff}} = 11$ layers and $d_{\text{ff}}/2 = 1024$ rotations per layer, this requires $11 \times 1024 \times 6 = 67,584$ FLOPs.
\end{itemize}

Total per expert: 1,129,984 FLOPs. For top-$k=2$: 2,259,968 FLOPs/token.

\textbf{Memory Traffic:} DRAM traffic consists of the 202.2 KB shared substrate, 27.1 KB of FP16 butterfly angles per active expert and 5.0 KB for input and output tensors. In steady-state (substrate cached in L2), this requires only 58.0 KB of memory traffic per token. In a cold-cache scenario (first token), loading the shared substrate increases this to 260.2 KB. This is computed as
\begin{itemize}
    \item \textbf{Steady-state} ($W_{\text{base}}$ cached): Only angles and tensors are loaded per token.
    \begin{equation}
    \text{Traffic} = 1,024 + 2 \times 27,136 + 4,096 = 59,392 \text{ bytes} \approx 58.0 \text{ KB}
    \end{equation}
    \item \textbf{Cold cache} (first token): $W_{\text{base}}$ must also be read from DRAM.
    \begin{equation}
    \text{Traffic} = 207,094 + 59,392 = 266,486 \text{ bytes} \approx 260.2 \text{ KB}
    \end{equation}
\end{itemize}

\subsection{Arithmetic Intensity and Performance Bounds}
\begin{equation}
\text{AI}_{\text{cached}} = \frac{2,259,968}{59,392} = 38.1 \text{ FLOP/Byte}
\end{equation}
\begin{equation}
\text{AI}_{\text{cold}} = \frac{2,259,968}{266,486} = 8.5 \text{ FLOP/Byte}
\end{equation}
Both of these values exceed the ridge point of 8 FLOP/byte. Hence, ButterflyMoE is compute-bound on the Cortex-A72, achieving the full performance of 48 GFLOPS/second in both scenarios.

For comparison, Standard MoE and Dense FFN baselines load full FP16 weight matrices ($2 \times 2048 \times 512 \times 2 = 4.2$ MB per token for top-$k=2$), yielding:
\begin{equation}
\text{AI}_{\text{Standard MoE}} = \frac{4,194,304}{4,194,304} = 1.0 \text{ FLOP/Byte}
\end{equation}

This falls well below the ridge point, making both Standard MoE and Dense FFN memory-bound at only 6.0 GFLOP/s. Standard MoE and Dense FFN are memory bound at 12.5\% of peak compute.

\section{Conclusion}
ButterflyMoE dramatically reduces the per-expert memory cost in MoE architectures from $\mathcal{O}(d^2)$ to $\mathcal{O}(d \log d)$ through parameterization of experts as orbital transformations of a shared ternary substrate. This method asymptotically yields up to 150$\times$ 
compression at 256 experts. Learned per expert rotations enable quantization which suppress activation outliers, a problem that static rotation methods could not solve

ButterflyMoE reduces the memory bottleneck, enabling exploration of deployment scenarios at scale. 64-experts can theoretically fit in 1.9 MB and 256 experts can fit in 6.82 MB showing massive memory gains. While empirical on-device profiling is left to future work, our theoretical roofline analysis confirms that this memory reduction fundamentally alters the execution regime on representative edge processors whereas Standard MoE is still memory bound on edge processors. For edge devices with a memory budget, reducing per-expert cost by a factor of $\frac{d}{\log d}$ is not an optimization but a requirement.

 A key limitation of our current evaluation is the scale of the models, which is constrained by our available hardware, a single Tesla T4. Our experiments serve as a proof of concept for the orbital parameterization mechanism rather than a demonstration of production-scale training.
Future work will focus on two critical extensions:scaling to wider architectures with 64-128 experts to study routing granularity and moving our theoretical roof line model to perform empirical on-device profiling on edge processors to quantify real-world energy and latency improvements.
The entire code would be released upon acceptance.

\bibliographystyle{plain}
\bibliography{references}

\end{document}